\pdfoutput=1

\documentclass[11pt]{article}

\usepackage[]{acl}

\usepackage{times}
\usepackage{latexsym}
\usepackage{float}
\usepackage{graphicx}
\usepackage{multirow}
\usepackage{booktabs}
\usepackage{color}
\definecolor{c1}{RGB}{164,198,57}
\definecolor{c2}{RGB}{223, 113, 27}
\definecolor{c3}{RGB}{33,171,205}
\definecolor{c4}{RGB}{203,65,84}
\usepackage{bm}

\usepackage[T1]{fontenc}

\usepackage[utf8]{inputenc}

\usepackage{microtype}

\title{Draft, Command, and Edit: Controllable Text Editing in E-Commerce}
\author{Kexin Yang$^\spadesuit{\,}$~~~\textbf{Dayiheng Liu}$^\spadesuit{\,}$~~~\textbf{Wenqiang Lei}$^\spadesuit{\,}$~~~\\ \textbf{Baosong Yang}$^\diamondsuit$~~~\textbf{Qian Qu}$^\spadesuit{\,}$~~~ \textbf{Jiancheng Lv}$^\spadesuit{\,}$\\
$^\spadesuit$Sichuan University \\
$^\diamondsuit$University of Macau \\ 
\texttt{\{kexinyang0528, losinuris\}@gmail.com}
}
\begin{document}
\maketitle
\begin{abstract}
Product description generation is a challenging and under-explored task. Most such work takes a set of product attributes as inputs then generates a description from scratch in a single pass. However, this widespread paradigm might be limited when facing the dynamic wishes of users on constraining the description, such as deleting or adding the content of a user-specified attribute based on the previous version. To address this challenge, we explore a new \textit{draft-command-edit} manner in description generation, leading to the proposed new task---controllable text editing in E-commerce. More specifically, we allow systems to receive a command (deleting or adding) from the user and then generate a description by flexibly modifying the content based on the previous version. It is easier and more practical to meet the new needs by modifying previous versions than generating from scratch. Furthermore, we design a data augmentation method to remedy the low resource challenge in this task, which contains a model-based and a rule-based strategy to imitate the edit by humans. To accompany this new task, we present a human-written \textit{draft-command-edit} dataset called E-cEdits and a new metric ``Attribute Edit''. Our experimental results show that using the new data augmentation method outperforms baselines to a greater extent in both automatic and human evaluations.
\end{abstract}

\section{Introduction}
\begin{figure}[t]
   \centering
   \includegraphics[scale=0.5]{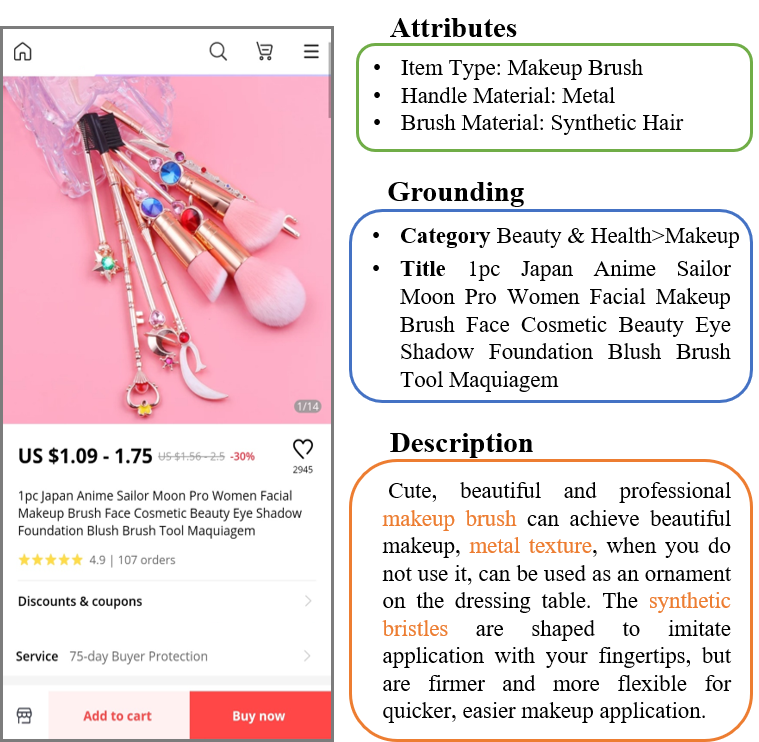}
   \caption{An example in our data source. We collect right data items from the left website. The attribute-relevant contents are colored in the description.}
   \label{fig:figure1}
\end{figure}

In E-commerce, controllable text generation plays an essential role in generating attractive and suitable product descriptions~\cite{controllable_divers_generation_e-commerce}. These automatic description generation methods bring significant increases in writing efficiency and cost savings when facing billions of product data~\cite{pattern_e-commerce}.

Most recent works depend on the single-pass paradigm to design the description generation manner~\cite{chinese_e-commerce,Stick_to_facts_e-commerce,controllable_divers_generation_e-commerce}. In details, the generation model takes a set of inputs, which include the key selling points of a product (i.e., a set of attributes~\cite{controllable_divers_generation_e-commerce}) and various forms of grounding, such as titles~\cite{pattern_e-commerce}, customer reviews~\cite{Probing_Product_Description_e-commerce}, or knowledge base~\cite{knowledge-based_e-commerce}. Then, it outputs the final description, without considering user feedback (i.e., in a single pass). 

Despite the success of the above studies, there is still a major limitation in this single-pass paradigm --- it fails to interact with users and flexibly refine the generated description. Specifically, users' expectations about the content might change after receiving the first version of description. For example, they may want to add or delete the content of a specific product attribute to obtain a more appropriate description. Unfortunately, it is inflexible for the single-pass paradigm to address this situation. On the one hand, existing models have to regenerate a description from scratch, even if needing a very few words changes from the previous version. On the other hand, utilizing manual post processing is time-consuming and often prohibitively expensive~\cite{manual_post_editing}, since the edit operation includes finding the right place from the entire description, then rewriting the attribute-relevant content while fine-tuning context to keep readability.

To reach the ideal goal of generating descriptions interactively, we propose a new task to approximate the condition---controllable E-commerce description editing. In short, we allow users to flexibly modify the previous description (hereinafter known as \textit{draft}) in a \textit{draft-command-edit} manner. In particular, users input a specific \textit{command} with an attribute (e.g., deleting or adding the attribute-relevant content in \textit{draft}), the model then generates a new description (i.e., \textit{edit}) based on the \textit{command}, \textit{draft}, and grounding (e.g., product title, product property). This paradigm would be easier and more practical to meet the new needs of users~\cite{Text_Editing_Command}, while making the first attempt to high-volume processing on user-oriented description editing. 

The key challenge of this task is that the rich supply of alignment data between \textit{draft} and \textit{edit} would be naturally inaccessible, which may cause big difficulties for these data-driven generation models. To overcome this low resource limitation, we propose a data augmentation method to automatically generate \textit{draft}-\textit{edit} data pairs. In more detail, we design two strategies to imitate the edit by humans. One of them uses a filling-in-the-blank generation model to approximate the human edit that removes the attribute-relevant tokens and slightly modifies the context to keep readability. Another one is based on the rules to imitate the edit that directly deletes the attribute tokens in the content.

Besides, we introduce a new \textit{draft-command-edit} dataset called \textbf{E-cEdits} and a novel metric ``Attribute  Edit'' to evaluate our method. E-cEdits is created by humans via crowdsourcing in Anonymity E-commerce scenario.\footnote{For the anonymous submission, we have temporarily hidden the specific real-world E-commerce platform name.} As Figure \ref{fig:figure1} shown, we first crawl data from the platform, then the human annotators are asked to edit the description until it excludes the content about a pre-specified attribute. In the end, there are 9,000 <attribute, command, grounding, draft, edit> 5-tuples from 733 product categories. In addition, ``Attribute Edit'' aims to examine whether an attribute has been edited. In details, it computes a fuzzy matching score between the input attribute and the model output, evaluating whether the content of the user-specified attribute appears (for adding commands) / disappears (for deleting commands) in final results. It is simple but effective, and our experiments demonstrate that ``Attribute Edit'' metric is significantly correlated with human evaluation. 


The main contributions of this work could be summarized as follows:
\begin{itemize}
    \item We propose a challenging new text generation task in E-commerce, which allows the model to flexibly modify the description constrained by users' dynamical requirements. This paradigm could provide a novel insight to design a user-oriented generation manner in controllable text generation. 
    
    \item Responding to the key challenge of low resource in this task, we propose a new automatic data augmentation method to approximate human edit, which automatically generates pseudo data. Experiments on the E-cEdits dataset show that our system significantly outperforms baselines in automatic and human evaluations.
    
    \item  To accompany this task, we release an E-commerce text editing dataset E-cEdits and design a novel metric ``Attribute Edit''.
    
\end{itemize}

\begin{table*}[t]
\small
\centering
\begin{tabular}{p{4.1cm}p{5.5cm}p{5cm}}
\toprule
Attribute & Description $\hat{x}$ & Description ${x}$  \\
\midrule
shape: \textcolor{c2}{column} & brass magnetic clasps, \textcolor{c2}{column}, silver size:about 8mm wide, ..., just add to the end of your diy bracelets crimp in the hole. & brass magnetic clasps, silver size:about 8mm wide, ..., just add to the end of your diy bracelets crimp in the hole.\\ \hline\hline

flavor: \textcolor{c1}{green-apple flavor} & you will receive \textcolor{c1}{green-apple flavor} a must have for braces, ..., package including : 10 boxes ortho wax. & you will receive a must have for braces, ..., package including : 10 boxes ortho wax. \\ \hline\hline

club type: \textcolor{c3}{hybrids} & brand new aftermarket adapter \textcolor{c3}{for taylormade hybrids}, ..., they have 1.5 degrees of loft adjustment. & brand new aftermarket adapter, ..., they have 1.5 degrees of loft adjustment. \\ \hline\hline

nintendo model: \textcolor{c4}{nintendo switch}	& charging data cable for \textcolor{c4}{nintendo switch} type c usb charger, ..., 1pcs x charging cable. & charging data cable for the one that uses type c usb charger, ..., 1pcs x charging cable.\\
\bottomrule
\end{tabular}
\caption{Example description edits of E-cEdits. The unmodified contents are omitted and the modified properties are colored. Best viewed in color.}
\label{table:example}
\end{table*}

\section{Related Work}
In this section, we first analyze the related works on text generation in E-commerce. Then, we review some representative works which introduce text editing into some text generation scenarios.

\subsection{Text Generation in E-commerce}
Recently, various attempts have been made in E-commerce text generation, such as title generation~\cite{e-commence_title_mane2020product}, summarization~\cite{e-commence_summarization_li2020aspect}, dialogue~\cite{e-commerce_dialogue_zhang2020multi}, and answer generation~\cite{e-commerce_answer_generation_gao2021meaningful}. The most related task to us is product description generation. Apex~\cite{controllable_divers_generation_e-commerce} based on a Conditional Variational Autoencoder generates a description from a set of attributes. FPDG~\cite{Stick_to_facts_e-commerce} 
considers the entity label of each word and increases the fidelity of the descriptions. KOBE~\cite{knowledge-based_e-commerce} takes a variety of factors into account while generating descriptions, including product aspects, user categories, and knowledge base. The previous methods have their applicative advantages and insurmountable disadvantages: for example, they consider the generation of description under the one-pass setting from scratch. However, the users' needs for constraining the description could be dynamic. In contrast, we provide a new generation paradigm regarding the process in a \textit{draft-command-edit} manner, which significantly drops the difficulty of the description generating by taking advantage of previous versions.

\subsection{Text Editing}
Dating back to the period of rule-based postediting~\cite{knight1994automated}, text editing has long been investigated for text generation. 
According to the differences in the ultimate goal, it can be roughly divided into two types: (1) Refining the sentences to be more fluency and factually grounded; (2) Adding or modifying the content. The first type has a set of different settings, such as post-editing~\cite{post_editing_machine, post_editing_FELIX}, grammatical error correction~\cite{error_correction_zhao2020maskgec,error_correction_wan2020improving}, and paraphrasing~\cite{paraphrasing_DBLP:conf/acl/GoyalD20,paraphrasing_siddique2020unsupervised}. Our new task is more relevant to the second type, in which editing text based on prototypes has achieved promising performance. For example, \citet{guu2018generating} sample prototype from the training corpus. FACTEDITOR~\cite{Text_Editing_Fact} creates a fact-based draft by rules as the model input in two data-to-text tasks. \citet{Text_Editing_Command} crawl data from Wikipedia's revision histories to form a Draft-Edit pair, and generate text according to the command in a progressively adding manner. However, directly incorporating the methods above into E-commerce is less portable. It is because the text editing manner of these methods is based on continuing writing (i.e., generating new sentences after the current text), which might be hard to modify previous content according to users' wishes. In comparison, our task provides more flexible operations---adding and deleting, which takes modifying and adding content of the description both into account. Meanwhile, the editing object product attribute is the central theme of a product description, which promotes our task more adaptable to the application requirements in E-commerce~\cite{attribute_important}.

\section{Preliminaries}
In this section, we introduce our new task and elaborate the benchmark E-cEdits. To ease of presentation, we start from formalize the new task in \S~\ref{sec:task}. Then we give a detailed creation of the presented dataset E-cEdits in \S~\ref{sec:dataset}. 

\subsection{Task Definition} \label{sec:task}
The controllable E-commerce description editing task is defined as follows: given an attribute $a$, a command $\mu$ and the specified forms of grounding $g$, the system generates a new description $x$ based on the previous version $\hat{x}$. In our specific settings to instantiate this task, the command $\mu$ is defined as adding or deleting the content of attribute $a$ from the description $\hat{x}$ while keeping readability. Meanwhile, the type of attribute-relevant content is various. It could be a word, a phrase, or a clause when appearing in the description. In addition, grounding is the supplementary information about the product, such as titles. In sum, the editing process is $\hat{x} \rightarrow x$, given $a$, $\mu$, and $g$. 

\subsection{E-cEdits} \label{sec:dataset}
It is tough to obtain $\hat{x}$ for $x$ from the E-commerce platform, since the history version for attributes editing is difficult to collect. To accompany this task and evaluate the effectiveness of the proposed method, we present a high-quality dataset containing 9,000 \textit{draft-command-edit} tuples in the English E-commerce domain, which is wholly written by humans via crowdsourcing in the Anonymity E-commerce platform. To be concrete, we ask each annotator to remove the content of a pre-specified attribute from the complete description, which is consistent with the deleting editing in the task definition.\footnote{In fact, each description has multiple attributes (2.61 on average), and we randomly select one of them as the pre-specified hint for description editing.} When considering the adding editing, data tuples can be easily obtained by exchanging the deleted editing samples' source and target descriptions. It is worth mentioning that the editing operation of humans follows the ``minimum modification principle''. This principle means that the annotator should remove the relevant content, may add punctuation or a few words to keep the readability. In addition, the third-party inspectors examine 200 random samples from the editing data for data quality assurance and make sure the pass rate is more than 95\%. Finally, each sample of E-cEdits contains 5-tuple <attribute, command, grounding, draft, edit>. In the implementation, the command includes two signals, ``[ADD]'' and ``[DEL]'', which denote adding and deleting commands, respectively. Meanwhile, we choose product title and category as the grounding following~\citet{pattern_e-commerce}.

\begin{table}[t]
\small
\centering
\begin{tabular}{llll}
\toprule
Statistic & Item & Numbers & Mean Length \\
\midrule
\multirow{2}*{Description} & $\hat{x}$ & 9,000 & 69.47\\
~ & $x$ & 9,000 & 65.87\\
\midrule
 \multirow{2}*{Grounding}& Category &  733 & - \\
~ & Title & 9,000 & 15.92\\
\midrule
Attribute & Attribute & 9,000\footnotemark[1] & 3.17\\
\bottomrule
\end{tabular}
\caption{Summary statistics of E-cEdits.}
\label{table:statistics}
\end{table}

\noindent\textbf{Statistical Analysis} Table~\ref{table:statistics} gives a statistical overview of our dataset. More concretely, the mean length difference between \textit{draft} (i.e., $\hat{x}$) and \textit{edit} (i.e., $x$) suggests that the words changing in editing is slightly. Meanwhile, descriptions come from 733 categories, and various types mean our dataset could approximate the real condition. Finally, the mean length of the attribute implies that it usually includes only a single word or a short phrase. As a result, the minimal information offered by attributes may bring algorithms difficulties to generate a description from scratch. 

\begin{figure*}[t]
   \centering
   \includegraphics[scale=0.45]{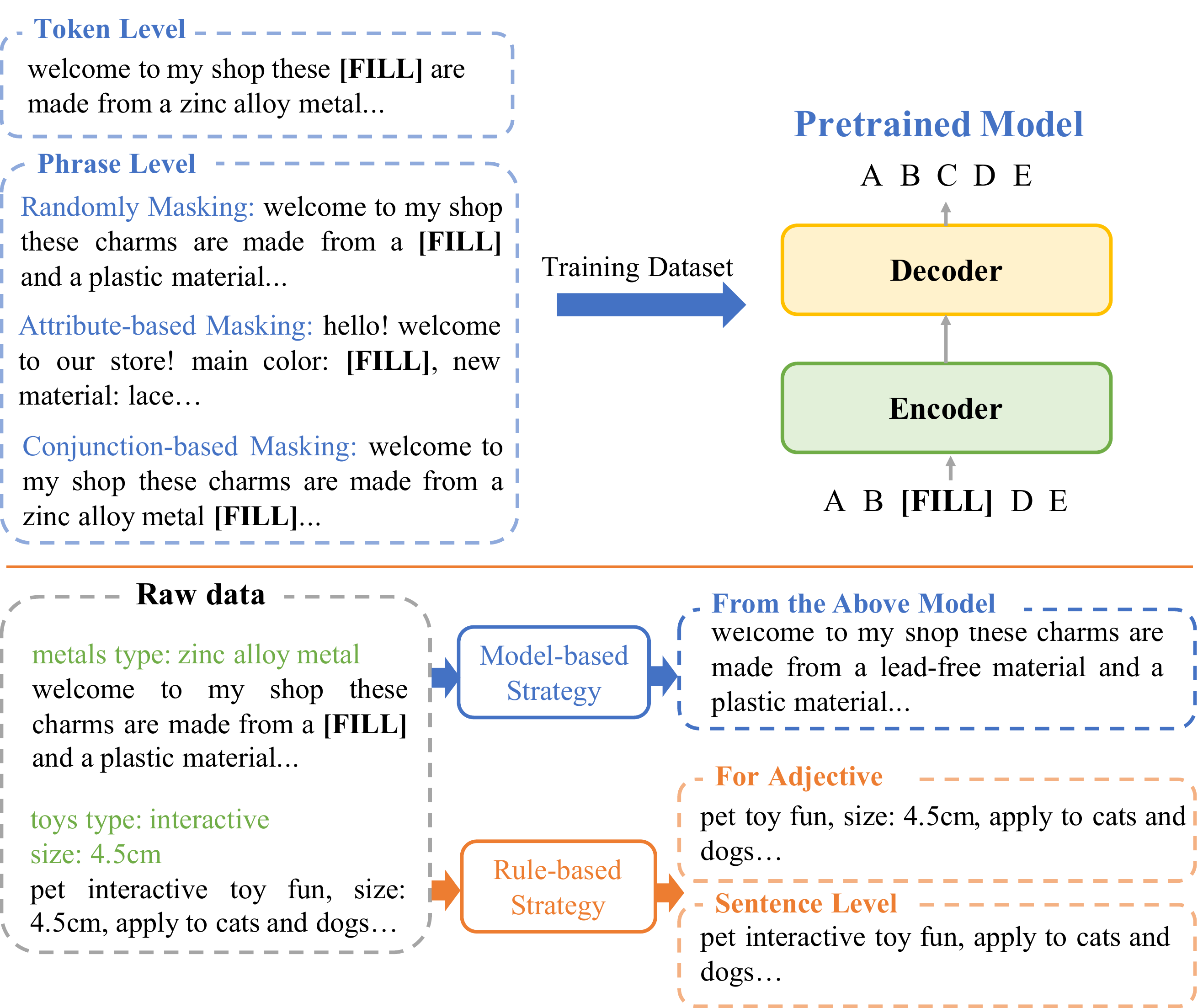}
   \caption{An overview of our data augmentation method. The upper part (orange line above) illustrates the model training stage of our model-based strategy.}
   \label{fig:figure2}
\end{figure*}

\noindent\textbf{Examples} The typical edit examples of E-cEdits are illustrated in Table~\ref{table:example}. The phenomena can be mainly divided into two categories: 1) Deleting the attribute-relevant content. For instance, the annotator may delete the keywords (row 2), the phrase (row 3), or the clause (row 1); 2) Replacing the attribute words with attribute-free ones (row 4) to keep the text flowing. We can see that the editing contains various forms and appears in different positions, which will present the editing system in the low-resource settings with a great challenge.

\section{Method}
In this section, we elaborate our data augmentation method and the description editing model. To ease of presentation, we start from toy examples to illustrate the overview of the data augmentation method in \S~\ref{sec:imp}, and give a detailed explanation of the implementation. Then, we present the editing model in \S~\ref{sec:training}.

\subsection{Automatic Data Augmentation} \label{sec:imp}
To remedy the low resource challenge in this task, we design a data augmentation method to imitate the edit by humans. Although \textit{draft-command-edit} data pairs are difficult to obtain naturally, a large number of descriptions and corresponding attributes can be easily collected from E-commerce platforms. Thus, we consider strategically removing the content of a pre-specified attribute from the description (i.e., the deleting editing), which is similar to the dataset building method of humans editing. After that, we obtain the adding editing sample by exchanging the source and target descriptions in deleting, as mentioned in \S~\ref{sec:dataset}.

\noindent\textbf{Model-based Strategy} Our basic idea is to mask the content of a pre-specified attribute in description with a signal ``\textbf{[FILL]}'' then use a filling-in-the-blank model to generate attribute-free content on that position. Therefore, we get a \textit{draft} for each description while keeping readability. The workflow is shown in the blue flow of Figure~\ref{fig:figure2}. For each description, we extract a word or a phrase, then replace it with a mask token ``\textbf{[FILL]}''. After that, we fine-tune a pre-trained Seq2Seq model (e.g., ProphetNet~\cite{ProphetNet}) to reconstruct the description (i.e., the ``Pretrained model'' part in Figure~\ref{fig:figure2}). In the inference, we use a phrase fuzzy matching tool based on the Levenshtein Distance to mask the content of a pre-specified attribute with ``\textbf{[FILL]}''.\footnote{This tool can be accessed via \url{https://pypi.org/project/fuzzywuzzy/}.} Then, we constrain the decoding space by removing the attribute tokens in vocabulary while generating. As a result, the model can generate an attribute-free description.

It is worth mentioning that we develop multiple policies for deciding which tokens or phrases should be masked with ``\textbf{[FILL]}''. First, we use the TF-IDF score at the token level to choose and mask an essential word in each description (except for stop words and punctuation). The TF-IDF score could provide the uniqueness and local importance of a word at the corpus level~\cite{POINTER}. Second, we design three masking approaches to increase the filling types' variety at the phrase level, aiming to approximate human editing situations. Concretely, it includes random, conjunction-based, and attribute-based masking (the blue block ``Phrase Level''). In randomly masking, we randomly choose 2-5 word pieces to mask each description, referring to the length statistic results. Meanwhile, we consider masking attribute-relevant words with phrase fuzzy matching tool thus models can enjoy benefits to better deal with attributes. In addition, we randomly mask the clauses connected with coordinating conjunctions as it is a typical form when multiple attributes appear in one sentence. Finally, we collect 1.2 million data pairs in total for model training. The proportion of each type in the training data set is: 50\% for token level and 50\% for phrase-level (50\% attribute-based masking, 30\% conjunction-based masking, and 20\% randomly masking). 

\noindent\textbf{Rule-based Strategy} We design two ways to imitate the editing type of directly deleting attribute-relevant content in a description. On the one hand, we directly remove the sentence in a description if it only contains one attribute (the ``Sentence Level'' block in the figure). On the other hand, as removing adjectives does not affect the sentence integrity, we also fall the attributes of adjectives into this category (the ``For adjective'' block).

\subsection{Model} \label{sec:training}
For the description edit task, we use the standard auto-regressive sequence to sequence models as test beds, thus various generation models can be easily adapted to this task. Given an attribute $a$, an command $\mu$, groundings $g$, and \textit{draft} $\hat{x}$, the model generates Edit $x=({\rm x}_{1},{\rm x}_{2},...,{\rm x}_{T})$ by:
\begin{equation}
    p( x|\hat{x},a,\mu,g;\theta)=\prod_{t=1}^{T}p({\rm x}_{t}|{\rm x}_{1:t-1},\hat{x},a,\mu,g;\theta),
\end{equation}
where $\theta$ is the model parameters. 

\begin{figure}[t]
   \centering
   \includegraphics[scale=0.48]{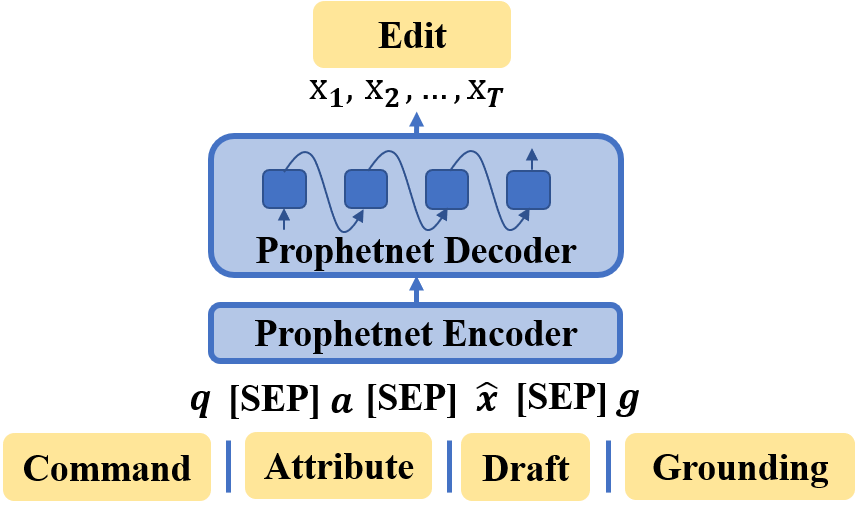}
   \caption{An overview of our edit model.}
   \label{fig:figure3}
\end{figure}

In implementation, we use the pre-trained Seq2Seq model ProphetNet~\cite{ProphetNet} as the backbone, which is effective on several text generation tasks~\cite{liu2020glge,gcpg}. To adapt the ProphetNet in our task, we follow~\citet{pretraining} to continue training it in three steps. Firstly, to adapt ProphetNet to the E-commerce domain, we collect 24 million grounding E-commerce titles to continue pre-train ProphetNet in a denoising sequence-to-sequence task. After that, we obtain a pre-trained E-commerce domain Seq2Seq model called ProphetNet-E. Secondly, we train ProphetNet-E in text editing task using our automatic augmentation dataset as described in \S~\ref{sec:imp}, which contains 600 thousand samples. Finally, we train the model on the E-cEdits dataset. As shown in Figure~\ref{fig:figure3}, in the editing task, it has to be mentioned that we concatenate all of the model's input ($a$, $\mu$, $g$, and $\hat{x}$) by the separator signal ``\textbf{[SEP]}'' to adapt the models for our task.


\section{Experiments}

\begin{table*}[t]
    \centering
    \small
    \begin{tabular}{lllllllll}
    \toprule
    Item & Model & \multicolumn{3}{c}{Attribute Edit} & 
     \multicolumn{3}{c}{ROUGE} & BLEU \\
     \cmidrule(lr){3-5} \cmidrule(lr){6-8}
        &   & ADD~$\uparrow$ & DEL~$\downarrow$ & ALL~$\uparrow$ & R-1~$\uparrow$ & R-2~$\uparrow$ & R-L~$\uparrow$ & B-4~$\uparrow$ \\
    \midrule
    \multirow{4}*{Baselines (row 1-3)}
    & Transformer  & 56.32 & 55.57 & 0.37 &  14.23 &  2.54 &  13.05 & 3.50 \\
    ~ & MASS  &  58.47 &  90.00 &  -15.76 & 94.72 & 92.05 & 94.69 & 89.30 \\
    ~ & ProphetNet  & 59.33 & 87.60 &  -14.43 & 91.59 & 88.98 & 91.53 & 84.90 \\
    \midrule
    \multirow{3}*{Ablations (row 4-6)} & only Grounding  &  61.72  &  85.77 & -12.03 & 89.16  & 86.00  &  89.03 & 79.97 \\ 
    ~ & only Command & 61.08 & 87.24 & -13.08 & 92.44 & 89.38 & 92.34 & 85.21 \\ 
    ~ & no Data Augmentation  &  61.24 &  85.97 & -12.36 & 89.53 & 86.45 & 89.42 & 80.69\\   
    \midrule
    ~ & Our system & 87.29  & 58.09 & 14.60 & 96.52 & 94.01 & 96.28 & 91.78\\
    \bottomrule
    \end{tabular}
    \caption{Performances of our system and baselines on E-cEdits dataset in terms of Fluency (BLEU and ROUGE) and Attribute relevance (Attribute Edit). R-1, R-2, and R-L denote ROUGE-1, ROUGE-2, and ROUGE-L, respectively. B-4 represents four gram BLEU score. All of ablations are based on ProphetNet-E.}
    \label{tab:result}
\end{table*}
\begin{table*}[t]
    \centering
    \small
    \begin{tabular}{llllllll}
        \toprule
         Model & \multicolumn{3}{c}{Fluency~$\uparrow$ }& \multicolumn{3}{c}{Attribute-relevant~$\uparrow$} & Overall~$\uparrow$\\
         \cmidrule(lr){2-4} \cmidrule(lr){5-7}
         ~ & ADD & DEL & ALL & ADD & DEL & ALL & \\
         \midrule
          MASS & 3.61 & 3.75 & 3.68 & 2.11 & 1.99 & 2.05 & 2.87 \\
          Our system  & 3.93 & 3.95 & 3.94 & 4.08 & 3.82 & 3.95 & 3.95 \\
        \bottomrule   
    \end{tabular}
    \caption{Human Evaluation for Mass and our system. Note that the higher of ``DEL'' score represents the better quality as opposed to the automatic evaluation. The significance test is carried out for each group by the Two-Sample t-Test, all of the p values are less than 0.05.}
    \label{tab:human}
\end{table*}

\subsection{Experimental Settings}
For the E-cEdits dataset, we randomly sample 4,000/1,000 pairs for training/validation, and the remaining 4000 for testing. All of the models are implemented based on Fairseq~\cite{fairseq}, and the specific parameters setting for each model can be found in \S~\ref{sec:baseline}. We set the max training epoch to 10 for each model. In the inference step, the beam size and length penalty are set to 4 and 1.2 respectively, to calculate the main results without post-processing. 

\subsection{Baselines} \label{sec:baseline} 
\noindent\textbf{Transformer} is the most commonly used sequence to sequence model. We follow the hyperparameters of standard Transformer~\cite{transformer}.


\noindent\textbf{MASS}~\cite{mass} uses the encoder-decoder framework to reconstruct a sentence fragment given the remaining part of the sentence. We use the largest released pretraining model ``MASS-middle-uncased'' trained on Wikipekia and BookCorpus. It contains six layers (embedding/hidden size 1024 and 16 head for each attention layer) for both encoder and decoder.\footnote{\url{https://github.com/microsoft/MASS}} 

\noindent\textbf{ProphetNet}~\cite{ProphetNet} introduces an n-stream self-attention mechanism and a self-supervised objective named future n-gram prediction. Two versions of ProphetNet are used, and the main difference is the source of the training dataset. ``ProphetNet'' in Table~\ref{tab:result} is trained on English Wikipekia and BookCorpus (16GB in total), while ``ProphetNet-E'' is continue trained on E-commerce text. Both of two models include a 12-layer encoder (decoder) with 1024 embedding/hidden size and 4096 feed-forward filter size.\footnote{\url{https://github.com/microsoft/ProphetNet}}  


\subsection{Evaluation Metrics}

\noindent\textbf{Automatic Evaluation} Following the automatic evaluation methods in both text editing~\cite{Text_Editing_Command, Text_Editing_Fact} and description generation in E-commerce~\cite{knowledge-based_e-commerce,Stick_to_facts_e-commerce}, we first use BLEU~\cite{bleu} and ROUGE~\cite{rouge} to keep in line with previous works. The BLEU score is calculated by the built-in function in Fairseq, while the ROUGE score is calculated by the ``files2rouge'' tool.\footnote{\url{https://github.com/pltrdy/files2rouge}} 

While ROUGE and BLUE could evaluate fluency of the generated description, there is no way to explicitly examine whether an attribute has been edited, which is one of the key points in this task. To tackle this problem, we propose a new evaluation indicator ``Attribute Edit'', which computes the fuzzy matching score between the input attribute and the model output (the matching tool is introduced in \S~\ref{sec:imp}, and score range 0-100). It is worth mentioning that this indicator is significantly correlated with human evaluation (details can be found in \S~\ref{sec:further}). 
Note that we use ``ADD'' and ``DEL'' to represent the evaluation scores on adding commands and deleting commands, while ``ALL'' denotes the average score of them. In deleting operations, the lower matching score indicates better model performance. Therefore, we convert ``DEL'' scores into negative ones when computing the overall score (i.e., ``ALL'' in Table~\ref{tab:result}). 

\noindent\textbf{Human Evaluation} We also conduct a human evaluation to compare our system with baselines. 10 human graders are asked to evaluate the fluency and attribute relevance (for deleting command, we evaluate the attribute irrelevance) across 50 randomly selected examples from our test set, which include 25 deleting editing samples and 25 adding editing samples. Following Genie~\cite{genie}, we use a discrete Likert scale for 5 categories: exceptionally bad, bad, just OK, good, and perfect instead of a continuous one. Finally, we convert categories into scores (1-5 from exceptionally bad to perfect) and get the average score. 

\begin{table}[h]
    \centering
    \small
    \begin{tabular}{lll}
        \toprule
        Item & Attribute Edit & Character-level LD\\
        \midrule
        ADD & 0.92 (p-value<0.0001) & -0.01 (p-value=0.96) \\
        DEL & 0.89 (p-value<0.0001) & 0.17 (p-value=0.23) \\
        \bottomrule   
    \end{tabular}
    \caption{The coefficients of Pearson correlation with human evaluation. ``LD'' denotes Levenshtein distance.}
    \label{tab:person}
\end{table}

\subsection{Main Results}
We evaluate the performance of our system and baselines on E-cEdits dataset and further provide ablations. We report the main result on Table~\ref{tab:result} and the human evaluation can be found in Table~\ref{tab:human}, from which we can make the following conclusions:

\noindent\textbf{1. Our system consistently outperforms baselines both in automatic evaluation and human evaluation.} For automatic evaluation (as shown in Table~\ref{tab:result}), our system outperforms all baselines both in fluency (BLEU and ROUGE) and attribute relevance (Attribute Edit). For example, comparing with MASS that gets the highest 89.30 BLUE and 94.72/92.05/94.69 ROUGE scores in text editing baselines (row 2), our system still beats MASS with 91.78 BLEU and 96.52/94.01/96.28 ROUGE score (row 7). Especially in Attribute Edit, using the proposed data augmentation method can significantly improve the editing model compared with other baselines.
We choose MASS to compare with our system in human evaluations because it is the best baseline in both ROUGE and BLUE. As shown in Table~\ref{tab:human}, our system also outperforms MASS in both fluency and attribute relevance. Consistent with the automatic evaluation, these two models show significant differences in ``Attribute-relevant'' scores (2.05 vs. 3.95).

\begin{table*}[t]
    \centering
    \small
    \begin{tabular}{p{15cm}}
        \toprule
         \textbf{Attribute: } \textcolor{c1}{Chain style: handmade chains}\\
         \textbf{Command: } Adding\\
         \textbf{Draft: } Beads for necklaces bracelets making, with silver iron eye pins, mixed stone...\\
         \textbf{Model output: } \textcolor{c1}{Handmade stone chips beads chains for} necklaces bracelets making, with silver iron eye pins, mixed stone...\\
         \hline\hline
         \textbf{Attribute: } \textcolor{c3}{Item name: original rear camera module replacement part}\\
         \textbf{Command: } Adding\\
         \textbf{Draft: } For sony xperia xa ultra compatible: for sony xperia xa ultra (only fit for xa ultra, not fit for xa1 ultra) features: original high...  \\
         \textbf{Model output: } \textcolor{c3}{Original rear camera module replacement part for sony xperia xa ultra (only fit for xperia ultra)}, not fit for xa1 ultra, features: original high... \\  
         \hline\hline
         \textbf{Attribute:} \textcolor{c2}{Suitable: abs systems}\\
         \textbf{Command: } Deleting\\
         \textbf{Draft:} Feature : automatic brake... and magnetic reservoir base to prevent accidental spillage. \textcolor{c2}{Suitable for use with abs systems.} Supplied with range of... \\
         \textbf{Model output:} Feature : automatic brake... and magnetic reservoir base to prevent accidental spillage . Supplied with range... \\
         \hline\hline
         \textbf{Attribute: } \textcolor{c4}{Product type: stopper}\\
         \textbf{Command: } Deleting\\
         \textbf{Draft: } We have our own button factory... bulk order have more discount. \textcolor{c4}{Item description: plastic stoppers: 24mm width * 23mm height.}\\
         \textbf{Model output: } We have our own button factory... bulk order have more discount.\\
        \bottomrule   
    \end{tabular}
    \caption{Examples for adding and deleting in our final results, where the colored words are matched with attributes.}
    \label{tab:case}
\end{table*}

\noindent\textbf{2. Both command and grounding play important roles in the controllable text editing task.} We also explore the impact of grounding and command on model performance. As shown in ablations (row 4 and row 5), using explicit command may improve performance in fluency, while using the grounding benefits model in better attribute editing. Meanwhile, ProphetNet-E outperforms all baselines (row 6) when in the ``no Data Augmentation'' condition. That is, using E-commerce text for pretraining could better adapt ProphetNet to this task (compared with row 3). 

\subsection{Further Analysis} \label{sec:further}
\noindent\textbf{Attribute Relevance Evaluation} We further compute Pearson correlation coefficient between the human score and our ``Attribute Edit'' score, to verify it can effectively evaluate whether models carry out an editing operation. Meanwhile, we also choose the character-level Levenshtein distance as the baseline, which is widely used in judging the two sentences’ similarity~\cite{distance}.\footnote{\url{https://pypi.org/project/python-Levenshtein/0.11.2/}}  Table~\ref{tab:person} illustrates that there is a significant statistical correlation between the proposed ``Attribute Edit'' score and human evaluation. In addition, the character-level Levenshtein distance is irrelevant with human evaluation as all of the p-values greater than 0.05.

\noindent\textbf{Case Analyze} Table~\ref{tab:case} illustrates four examples of description editing results with our edit system. Especially, we can see that the operation of adding and deleting is not just simply copying or removing all the words in the pre-specified attribute, as finding an appropriate position in \textit{draft} for editing operation is one of the challenges in this task. For example, with deleting command, sample 4 needs to remove the attribute ``product type: stopper''. Our model not only removes the word ``stopper'', but also the relevant content ``24mm width...'', which keeps the readability of the Edit version.    

\section{Conclusion}
In this paper, we propose a new controllable text editing task allowing users flexibility to constrain the attribute-relevant content of the product description by commands in a \textit{draft-command-edit} manner, and introduce a high-quality \textit{draft-command-edit} dataset E-cEdits written by humans. Meanwhile, in response to the low resource condition---the key challenge in this task, we design a data augmentation method that contains two strategies to generate pseud data pairs. Experiments demonstrate that our method significantly and consistently outperforms baselines both in automatic evaluation and human evaluation. In sum, as a new attempt, we tentatively give a simple but effective implementation of product description editing, successfully approximate the ideal goal of generating descriptions interactively. Thus in the future, such a paradigm deserves a closer and more detailed exploration. Therefore, we will investigate to design this interactive generation manner in a more superior way.

\bibliography{ref}

\begin{thebibliography}{34}
\expandafter\ifx\csname natexlab\endcsname\relax\def\natexlab#1{#1}\fi

\bibitem[{Chan et~al.(2019)Chan, Chen, Wang, Li, Zhang, Gai, Zhao, and
  Yan}]{Stick_to_facts_e-commerce}
Zhangming Chan, Xiuying Chen, Yongliang Wang, Juntao Li, Zhiqiang Zhang, Kun
  Gai, Dongyan Zhao, and Rui Yan. 2019.
\newblock Stick to the facts: Learning towards a fidelity-oriented e-commerce
  product description generation.
\newblock In \emph{Proceedings of the 2019 Conference on Empirical Methods in
  Natural Language Processing and the 9th International Joint Conference on
  Natural Language Processing (EMNLP-IJCNLP)}, pages 4960--4969.

\bibitem[{Chen et~al.(2019)Chen, Lin, Zhang, Yang, Zhou, and
  Tang}]{knowledge-based_e-commerce}
Qibin Chen, Junyang Lin, Yichang Zhang, Hongxia Yang, Jingren Zhou, and Jie
  Tang. 2019.
\newblock Towards knowledge-based personalized product description generation
  in e-commerce.
\newblock In \emph{Proceedings of the 25th ACM SIGKDD International Conference
  on Knowledge Discovery \& Data Mining}, pages 3040--3050.

\bibitem[{Faltings et~al.(2020)Faltings, Galley, Hintz, Brockett, Quirk, Gao,
  and Dolan}]{Text_Editing_Command}
Felix Faltings, Michel Galley, Gerold Hintz, Chris Brockett, Chris Quirk,
  Jianfeng Gao, and Bill Dolan. 2020.
\newblock Text editing by command.
\newblock \emph{arXiv preprint arXiv:2010.12826}.

\bibitem[{Gao et~al.(2021)Gao, Chen, Ren, Zhao, and
  Yan}]{e-commerce_answer_generation_gao2021meaningful}
Shen Gao, Xiuying Chen, Zhaochun Ren, Dongyan Zhao, and Rui Yan. 2021.
\newblock Meaningful answer generation of e-commerce question-answering.
\newblock \emph{ACM Transactions on Information Systems (TOIS)}, 39(2):1--26.

\bibitem[{Goyal and Durrett(2020)}]{paraphrasing_DBLP:conf/acl/GoyalD20}
Tanya Goyal and Greg Durrett. 2020.
\newblock Neural syntactic preordering for controlled paraphrase generation.
\newblock In \emph{Proceedings of the 58th Annual Meeting of the Association
  for Computational Linguistics, {ACL} 2020, Online, July 5-10, 2020}, pages
  238--252. Association for Computational Linguistics.

\bibitem[{Green et~al.(2013)Green, Heer, and Manning}]{manual_post_editing}
Spence Green, Jeffrey Heer, and Christopher~D Manning. 2013.
\newblock The efficacy of human post-editing for language translation.
\newblock In \emph{Proceedings of the SIGCHI conference on human factors in
  computing systems}, pages 439--448.

\bibitem[{Gururangan et~al.(2020)Gururangan, Marasovic, Swayamdipta, Lo,
  Beltagy, Downey, and Smith}]{pretraining}
Suchin Gururangan, Ana Marasovic, Swabha Swayamdipta, Kyle Lo, Iz~Beltagy, Doug
  Downey, and Noah~A. Smith. 2020.
\newblock \href {https://doi.org/10.18653/v1/2020.acl-main.740} {Don't stop
  pretraining: Adapt language models to domains and tasks}.
\newblock In \emph{Proceedings of the 58th Annual Meeting of the Association
  for Computational Linguistics, {ACL} 2020, Online, July 5-10, 2020}, pages
  8342--8360. Association for Computational Linguistics.

\bibitem[{Guu et~al.(2018)Guu, Hashimoto, Oren, and Liang}]{guu2018generating}
Kelvin Guu, Tatsunori~B Hashimoto, Yonatan Oren, and Percy Liang. 2018.
\newblock prototypes generating sentences by editing prototypes.
\newblock \emph{Transactions of the Association for Computational Linguistics},
  6:437--450.

\bibitem[{Herbig et~al.(2020)Herbig, D{\"u}wel, Pal, Meladaki, Monshizadeh,
  Kr{\"u}ger, and van Genabith}]{post_editing_machine}
Nico Herbig, Tim D{\"u}wel, Santanu Pal, Kalliopi Meladaki, Mahsa Monshizadeh,
  Antonio Kr{\"u}ger, and Josef van Genabith. 2020.
\newblock Mmpe: A multi-modal interface for post-editing machine translation.
\newblock In \emph{Proceedings of the 58th Annual Meeting of the Association
  for Computational Linguistics}, pages 1691--1702.

\bibitem[{Iso et~al.(2020)Iso, Qiao, and Li}]{Text_Editing_Fact}
Hayate Iso, Chao Qiao, and Hang Li. 2020.
\newblock Fact-based text editing.
\newblock In \emph{Proceedings of the 58th Annual Meeting of the Association
  for Computational Linguistics, {ACL} 2020, Online, July 5-10, 2020}, pages
  171--182. Association for Computational Linguistics.

\bibitem[{Khashabi et~al.(2021)Khashabi, Stanovsky, Bragg, Lourie, Kasai, Choi,
  Smith, and Weld}]{genie}
Daniel Khashabi, Gabriel Stanovsky, Jonathan Bragg, Nicholas Lourie, Jungo
  Kasai, Yejin Choi, Noah~A Smith, and Daniel~S Weld. 2021.
\newblock Genie: A leaderboard for human-in-the-loop evaluation of text
  generation.
\newblock \emph{arXiv preprint arXiv:2101.06561}.

\bibitem[{Knight and Chander(1994)}]{knight1994automated}
Kevin Knight and Ishwar Chander. 1994.
\newblock Automated postediting of documents.
\newblock In \emph{AAAI}, volume~94, pages 779--784.

\bibitem[{Li et~al.(2020{\natexlab{a}})Li, Yuan, Xu, Wu, He, and
  Zhou}]{chinese_e-commerce}
Haoran Li, Peng Yuan, Song Xu, Youzheng Wu, Xiaodong He, and Bowen Zhou.
  2020{\natexlab{a}}.
\newblock Aspect-aware multimodal summarization for chinese e-commerce
  products.
\newblock In \emph{Proceedings of the AAAI Conference on Artificial
  Intelligence}, volume~34, pages 8188--8195.

\bibitem[{Li et~al.(2020{\natexlab{b}})Li, Yuan, Xu, Wu, He, and
  Zhou}]{e-commence_summarization_li2020aspect}
Haoran Li, Peng Yuan, Song Xu, Youzheng Wu, Xiaodong He, and Bowen Zhou.
  2020{\natexlab{b}}.
\newblock Aspect-aware multimodal summarization for chinese e-commerce
  products.
\newblock In \emph{Proceedings of the AAAI Conference on Artificial
  Intelligence}, volume~34, pages 8188--8195.

\bibitem[{Lin(2004)}]{rouge}
Chin-Yew Lin. 2004.
\newblock {ROUGE}: A package for automatic evaluation of summaries.
\newblock In \emph{Text Summarization Branches Out}, pages 74--81, Barcelona,
  Spain. Association for Computational Linguistics.

\bibitem[{Liu et~al.(2020)Liu, Yan, Gong, Qi, Zhang, Jiao, Chen, Fu, Shou, Gong
  et~al.}]{liu2020glge}
Dayiheng Liu, Yu~Yan, Yeyun Gong, Weizhen Qi, Hang Zhang, Jian Jiao, Weizhu
  Chen, Jie Fu, Linjun Shou, Ming Gong, et~al. 2020.
\newblock Glge: A new general language generation evaluation benchmark.
\newblock \emph{arXiv preprint arXiv:2011.11928}.

\bibitem[{Mallinson et~al.(2020)Mallinson, Severyn, Malmi, and
  Garrido}]{post_editing_FELIX}
Jonathan Mallinson, Aliaksei Severyn, Eric Malmi, and Guillermo Garrido. 2020.
\newblock {FELIX:} flexible text editing through tagging and insertion.
\newblock In \emph{Proceedings of the 2020 Conference on Empirical Methods in
  Natural Language Processing: Findings, {EMNLP} 2020, Online Event, 16-20
  November 2020}, pages 1244--1255. Association for Computational Linguistics.

\bibitem[{Mane et~al.(2020)Mane, Kedia, Mantha, Guo, and
  Achan}]{e-commence_title_mane2020product}
Mansi~Ranjit Mane, Shashank Kedia, Aditya Mantha, Stephen Guo, and Kannan
  Achan. 2020.
\newblock Product title generation for conversational systems using bert.
\newblock \emph{arXiv preprint arXiv:2007.11768}.

\bibitem[{Ott et~al.(2019)Ott, Edunov, Baevski, Fan, Gross, Ng, Grangier, and
  Auli}]{fairseq}
Myle Ott, Sergey Edunov, Alexei Baevski, Angela Fan, Sam Gross, Nathan Ng,
  David Grangier, and Michael Auli. 2019.
\newblock fairseq: A fast, extensible toolkit for sequence modeling.
\newblock In \emph{Proceedings of NAACL-HLT 2019: Demonstrations}.

\bibitem[{Papineni et~al.(2002)Papineni, Roukos, Ward, and Zhu}]{bleu}
Kishore Papineni, Salim Roukos, Todd Ward, and Wei{-}Jing Zhu. 2002.
\newblock Bleu: a method for automatic evaluation of machine translation.
\newblock In \emph{Proceedings of the 40th Annual Meeting of the Association
  for Computational Linguistics, July 6-12, 2002, Philadelphia, PA, {USA}},
  pages 311--318. {ACL}.

\bibitem[{Petrovski and Bizer(2017)}]{attribute_important}
Petar Petrovski and Christian Bizer. 2017.
\newblock Extracting attribute-value pairs from product specifications on the
  web.
\newblock In \emph{Proceedings of the International Conference on Web
  Intelligence}, pages 558--565.

\bibitem[{Qi et~al.(2020)Qi, Yan, Gong, Liu, Duan, Chen, Zhang, and
  Zhou}]{ProphetNet}
Weizhen Qi, Yu~Yan, Yeyun Gong, Dayiheng Liu, Nan Duan, Jiusheng Chen, Ruofei
  Zhang, and Ming Zhou. 2020.
\newblock Prophetnet: Predicting future n-gram for sequence-to-sequence
  pre-training.
\newblock In \emph{Proceedings of the 2020 Conference on Empirical Methods in
  Natural Language Processing: Findings}, pages 2401--2410.

\bibitem[{Shao et~al.(2021)Shao, Wang, Lin, Zhang, Zhang, Ji, and
  Abdelzaher}]{controllable_divers_generation_e-commerce}
Huajie Shao, Jun Wang, Haohong Lin, Xuezhou Zhang, Aston Zhang, Heng Ji, and
  Tarek~F. Abdelzaher. 2021.
\newblock Controllable and diverse text generation in e-commerce.
\newblock In \emph{The World Wide Web Conference}.

\bibitem[{Siddique et~al.(2020)Siddique, Oymak, and
  Hristidis}]{paraphrasing_siddique2020unsupervised}
AB~Siddique, Samet Oymak, and Vagelis Hristidis. 2020.
\newblock Unsupervised paraphrasing via deep reinforcement learning.
\newblock In \emph{Proceedings of the 26th ACM SIGKDD International Conference
  on Knowledge Discovery \& Data Mining}, pages 1800--1809.

\bibitem[{Snover et~al.(2006)Snover, Dorr, Schwartz, Micciulla, and
  Makhoul}]{distance}
Matthew Snover, Bonnie Dorr, Richard Schwartz, Linnea Micciulla, and John
  Makhoul. 2006.
\newblock A study of translation edit rate with targeted human annotation.
\newblock In \emph{Proceedings of association for machine translation in the
  Americas}, volume 200. Citeseer.

\bibitem[{Song et~al.(2019)Song, Tan, Qin, Lu, and Liu}]{mass}
Kaitao Song, Xu~Tan, Tao Qin, Jianfeng Lu, and Tie{-}Yan Liu. 2019.
\newblock \href {http://proceedings.mlr.press/v97/song19d.html} {{MASS:} masked
  sequence to sequence pre-training for language generation}.
\newblock In \emph{Proceedings of the 36th International Conference on Machine
  Learning, {ICML} 2019, 9-15 June 2019, Long Beach, California, {USA}},
  volume~97 of \emph{Proceedings of Machine Learning Research}, pages
  5926--5936. {PMLR}.

\bibitem[{Vaswani et~al.(2017)Vaswani, Shazeer, Parmar, Uszkoreit, Jones,
  Gomez, Kaiser, and Polosukhin}]{transformer}
Ashish Vaswani, Noam Shazeer, Niki Parmar, Jakob Uszkoreit, Llion Jones,
  Aidan~N. Gomez, Lukasz Kaiser, and Illia Polosukhin. 2017.
\newblock \href {http://papers.nips.cc/paper/7181-attention-is-all-you-need}
  {Attention is all you need}.
\newblock In \emph{Advances in Neural Information Processing Systems 30: Annual
  Conference on Neural Information Processing Systems 2017, December 4-9, 2017,
  Long Beach, CA, {USA}}, pages 5998--6008.

\bibitem[{Wan et~al.(2020)Wan, Wan, and
  Wang}]{error_correction_wan2020improving}
Zhaohong Wan, Xiaojun Wan, and Wenguang Wang. 2020.
\newblock Improving grammatical error correction with data augmentation by
  editing latent representation.
\newblock In \emph{Proceedings of the 28th International Conference on
  Computational Linguistics}, pages 2202--2212.

\bibitem[{Yang et~al.(2022)Yang, Liu, Lei, Yang, Zhang, Zhao, Yao, and
  Chen}]{gcpg}
Kexin Yang, Dayiheng Liu, Wenqiang Lei, Baosong Yang, Haibo Zhang, Xue Zhao,
  Wenqing Yao, and Boxing Chen. 2022.
\newblock \href {https://doi.org/10.18653/v1/2022.findings-acl.318} {{GCPG:}
  {A} general framework for controllable paraphrase generation}.
\newblock In \emph{Findings of the Association for Computational Linguistics:
  {ACL} 2022, Dublin, Ireland, May 22-27, 2022}, pages 4035--4047. Association
  for Computational Linguistics.

\bibitem[{Zhan et~al.(2021)Zhan, Zhang, Chen, Shen, Ding, Bao, Yan, and
  Lan}]{Probing_Product_Description_e-commerce}
Haolan Zhan, Hainan Zhang, Hongshen Chen, Lei Shen, Zhuoye Ding, Yongjun Bao,
  Weipeng Yan, and Yanyan Lan. 2021.
\newblock Probing product description generation via posterior distillation.

\bibitem[{Zhang et~al.(2019)Zhang, Zhang, Huo, and Ren}]{pattern_e-commerce}
Tao Zhang, Jin Zhang, Chengfu Huo, and Weijun Ren. 2019.
\newblock Automatic generation of pattern-controlled product description in
  e-commerce.
\newblock In \emph{The World Wide Web Conference}, pages 2355--2365.

\bibitem[{Zhang et~al.(2020{\natexlab{a}})Zhang, Song, Kang, Wang, Sun, Liu,
  Li, Zhang, and Si}]{e-commerce_dialogue_zhang2020multi}
WeiSheng Zhang, Kaisong Song, Yangyang Kang, Zhongqing Wang, Changlong Sun,
  Xiaozhong Liu, Shoushan Li, Min Zhang, and Luo Si. 2020{\natexlab{a}}.
\newblock Multi-turn dialogue generation in e-commerce platform with the
  context of historical dialogue.
\newblock In \emph{Proceedings of the 2020 Conference on Empirical Methods in
  Natural Language Processing: Findings}, pages 1981--1990.

\bibitem[{Zhang et~al.(2020{\natexlab{b}})Zhang, Wang, Li, Gan, Brockett, and
  Dolan}]{POINTER}
Yizhe Zhang, Guoyin Wang, Chunyuan Li, Zhe Gan, Chris Brockett, and Bill Dolan.
  2020{\natexlab{b}}.
\newblock \href {https://doi.org/10.18653/v1/2020.emnlp-main.698} {{POINTER:}
  constrained progressive text generation via insertion-based generative
  pre-training}.
\newblock In \emph{Proceedings of the 2020 Conference on Empirical Methods in
  Natural Language Processing, {EMNLP} 2020, Online, November 16-20, 2020},
  pages 8649--8670. Association for Computational Linguistics.

\bibitem[{Zhao and Wang(2020)}]{error_correction_zhao2020maskgec}
Zewei Zhao and Houfeng Wang. 2020.
\newblock Maskgec: Improving neural grammatical error correction via dynamic
  masking.
\newblock In \emph{Proceedings of the AAAI Conference on Artificial
  Intelligence}, volume~34, pages 1226--1233.

\end{thebibliography}
\bibliographystyle{acl_natbib}
\end{document}